\documentclass[conference]{IEEEtran}
\IEEEoverridecommandlockouts

\usepackage{cite}                      
\usepackage{amsmath,amssymb,amsfonts}  
\usepackage{algorithmic}               
\usepackage{graphicx}                  
\usepackage{textcomp}                  
\usepackage{xcolor}                    
\usepackage{url}                       
\usepackage{setspace}                  
\usepackage{tikz}                      
\usetikzlibrary{shapes, arrows, positioning, fit, calc, backgrounds}
\usepackage{booktabs}                  

\graphicspath{{figures/}{png/}}

\title{Hierarchical Deep Learning for Diatom Image Classification: A Multi-Level Taxonomic Approach}

\author{
\IEEEauthorblockN{Yueying Ke}
\IEEEauthorblockA{
\textit{Department of Computer Science}\\
\textit{The University of Texas at Austin}\\
yueying.dina.ke@utexas.edu 
}
}

\begin{document}

\maketitle

\begin{abstract}
Accurate taxonomic identification of diatoms is essential for aquatic ecosystem monitoring, yet conventional methods depend heavily on expert taxonomists. Recent deep learning approaches improve automation, but most treat diatom recognition as flat classification, predicting only one taxonomic rank. We investigate whether embedding taxonomic hierarchy into neural network architectures can improve both accuracy and error locality.

We introduce DiatomCascadeNet (H-COFGS), a hierarchical convolutional network with five cascaded heads that jointly predict class, order, family, genus, and species. Each head receives shared backbone features and probability distributions from higher levels, with binary masks restricting predictions to valid descendants during training and inference. Using a filtered dataset of 1,456 diatom images covering 82 species, we compare hierarchical and flat models under identical settings.

H-COFGS matches flat baselines at the species level (69.4\% accuracy) while outperforming at all upper taxonomic levels. When species predictions fail, errors remain taxonomically local: 92.5\% of misclassified species are correctly predicted at the genus level, versus 67.2\% for flat baselines. H-COFGS reduces mean taxonomic distance by 38.2\% (1.209 vs. 1.955).

Progressive training reveals bidirectional mechanisms: hierarchical constraint masks operate top-down to constrain prediction space, while gradients from fine-grained levels propagate bottom-up through the shared backbone, refining features. This improves class accuracy from 96.2\% to 99.5\% and yields 6-8\% gains at upper levels, producing more robust, interpretable, and biologically aligned predictions for multi-level taxonomic classification.
\end{abstract}



\section{Introduction}

\subsection{Background and Motivation}

Diatoms are unicellular microalgae with ornate silica cell walls, and shifts in their communities provide sensitive indicators of environmental change~\cite{round1990,kelly1995}. Reliable taxonomic identification is therefore central to water-quality assessment, ecological monitoring, and biodiversity studies~\cite{krammer2012,stevenson1999}. Yet the standard workflow still depends on expert taxonomists who inspect microscope images by hand, a process that is time-consuming, labor-intensive, and prone to inter-observer variability. As monitoring programs continue to scale up, demand for automated and trustworthy identification pipelines keeps growing.

Recent progress in deep learning has improved automated diatom classification~\cite{krizhevsky2012imagenet,he2016deep,simonyan2015very}. Convolutional neural networks (CNNs) with transfer learning from ImageNet~\cite{deng2009imagenet} reach high accuracy at individual taxonomic ranks~\cite{gonzalez2025,diatomnet2024}, learning discriminative visual features directly from microscope imagery and removing the need for hand-crafted descriptors~\cite{pan2010survey}. Almost all of these approaches, however, treat the task as flat classification: the network predicts a single rank, typically genus or species, and the upper-level labels are obtained by traversing the taxonomy tree. This setup is simple and efficient, but it leaves open a central question: can architectures that embed the taxonomic hierarchy deliver tangible benefits over single-level prediction with post-hoc tree lookup?

\subsection{Research Gap and Objectives}

Biological taxonomy arranges diatoms into a nested hierarchy: coarse levels (e.g., class, order) are distinguished by macro-scale traits such as valve symmetry and raphe structure, whereas fine levels (genus, species) depend on subtle micro-scale cues like striae density and valve outline~\cite{round1990}. Deep convolutional networks likewise progress from simple to complex features across layers~\cite{zeiler2014visualizing,szegedy2015going}. This parallel suggests that aligning different taxonomic ranks with different parts of the network may better match the structure of the task and help control how errors propagate.

In a flat model, the network predicts the species label once and derives all higher-level labels deterministically; if the species prediction is wrong, the taxonomic distance of the error depends entirely on which incorrect species is chosen, and there is no supervision that favors taxonomically local mistakes. Hierarchical models introduce an explicit top-down prediction process that constrains how errors spread. The process resembles traversing from the root to a target leaf by sequentially selecting branches: the model first makes easier, high-level decisions before committing to fine-grained species labels, whereas flat models must choose directly among all leaves in one step. Whether these architectural constraints actually yield benefits over a simpler flat design is the empirical question we investigate.

We evaluate two approaches:

\begin{enumerate}
    \item \textit{Flat baselines}: Standard CNNs that predict only the species label. Upper-level labels are obtained post hoc by traversing the taxonomy tree.

    \item \textit{Hierarchical models}: Multi-output CNNs with five cascaded heads (class, order, family, genus, and species). Each head receives the backbone features concatenated with the probability distributions from all higher levels, implementing feature fusion between visual and hierarchical cues. During training, we use teacher forcing with ground-truth masks, which creates bidirectional information flow: (1) bottom-up through gradient backpropagation as all five losses update the shared backbone, and (2) top-down through hierarchical masks that constrain each level's output space. During inference, we switch to greedy hierarchical prediction: the argmax at each level determines the mask for the next level.
\end{enumerate}

Experiments cover datasets ranging from 1,456 to 3,522 images, depending on the taxonomic depth and minimum-sample filters. We measure not only single-level accuracy but also error locality when fine-grained predictions fail.

\subsection{Contributions}

Our main contributions are:

\begin{enumerate}
    \item \textit{Architectural innovation}: A hierarchical CNN with cascaded heads that embeds taxonomic structure via feature fusion and masked constraints, enabling bidirectional information flow through teacher forcing and gradient backpropagation.

    \item \textit{Hierarchical vs.\ flat comparison}: A systematic comparison across all taxonomic levels showing that hierarchical models match species-level accuracy while substantially improving upper-level performance and error locality.
    
    \item \textit{Mechanistic analysis via progressive training}: A progressive sequence (F-C, H-CO, H-COF, H-COFG, and H-COFGS) that reveals how fine-grained supervision propagates backward to enhance coarse-level representations.
    
\end{enumerate}

\subsection{Paper Organization}
Section~II reviews related work on automated diatom classification and hierarchical machine learning. Section~III details the dataset, model architectures, and training procedures. Section~IV presents experimental results. Section~V discusses key findings, limitations, and broader implications. Section~VI concludes with a summary and directions for future work.

\section{Related Work}

\begin{figure*}[!t]
\centering
\includegraphics[width=\textwidth]{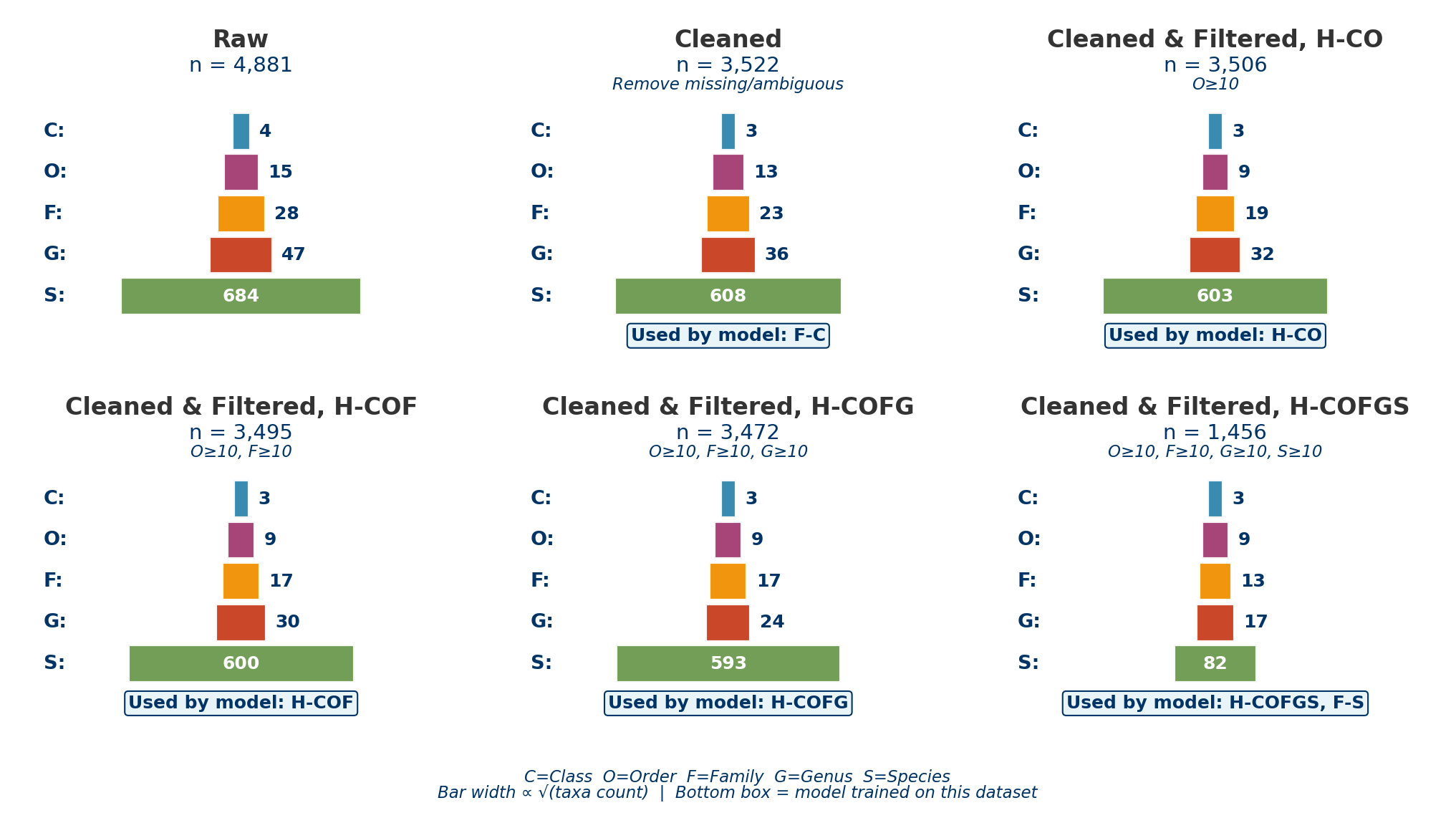}
\caption{Taxonomy pyramids across data pipeline stages. Each pyramid shows the number of taxonomic categories at each level (class, order, family, genus, and species). The pipeline progresses from raw data (4,881 samples) through taxonomic cleaning (3,522 samples) to progressively filtered datasets, enforcing minimum sample requirements.}
\label{fig:taxonomy_pyramids}
\end{figure*}
\subsection{Automated Diatom Classification}

Automated diatom identification has evolved significantly over the past two decades. Early approaches relied on hand-crafted features such as shape descriptors, texture analysis, and morphological measurements coupled with classical classifiers, including k-nearest neighbors, support vector machines, and decision trees \cite{pedraza2017,dimitrovski2012,chaushevska2020}. These methods encoded valve outline, striae density, and ornamentation into engineered feature vectors, which required substantial domain expertise and were sensitive to variations in focus, illumination, and specimen damage. Performance remained limited on large, taxonomically diverse datasets and did not scale well to hundreds of species.

The advent of deep learning has transformed diatom image analysis. Convolutional neural networks (CNNs), particularly in combination with transfer learning from large natural-image datasets such as ImageNet \cite{deng2009imagenet,krizhevsky2012imagenet}, have yielded substantial accuracy improvements over hand-crafted pipelines on both genus and species level tasks \cite{gonzalez2025,diatomnet2024}. Modern CNN architectures such as VGGNet, Inception, ResNet, and EfficientNet learn hierarchical representations automatically through successive convolutional and pooling layers \cite{simonyan2015very,szegedy2015going,he2016deep,tan2019efficientnet,zeiler2014visualizing}. Transfer learning leverages features learned on large-scale datasets to improve performance on specialized domains with limited training data \cite{yosinski2014transferable,pan2010survey,tan2018survey}.

Data augmentation techniques, including random rotations, flips, and color jittering, have been shown to improve generalization and reduce overfitting in image classification tasks \cite{shorten2019survey,krizhevsky2012imagenet}. Class imbalance, a common challenge in biological datasets, has been addressed through techniques such as focal loss, resampling strategies, and cost-sensitive learning \cite{lin2017focal,buda2018systematic,johnson2019survey}. However, most existing deep learning approaches treat diatom classification as a flat, single-level prediction problem, training models to predict a single taxonomic rank (typically genus or species) without explicitly modeling the underlying taxonomic hierarchy.

\subsection{Hierarchical Classification Methods}

Hierarchical classification is a well-established problem in machine learning, with applications spanning text categorization, image recognition, and biological taxonomy \cite{silla2011}. Silla and Freitas provide a comprehensive survey that distinguishes top-down strategies, which sequentially predict labels from root to leaf, from big-bang approaches that learn a single model to predict all levels jointly \cite{silla2011}. Early deep learning approaches include HD-CNN \cite{yan2015hdcnn}, which uses a coarse-to-fine cascade of CNNs, and hierarchical multi-label networks \cite{wehrmann2018hierarchical} that jointly predict multiple levels through shared representations.

Beyond simple decomposition strategies, subsequent work has introduced hierarchical regularization schemes that encode ancestor-descendant relations directly into the loss function, so that misclassifications near the true node are penalized less severely than taxonomically distant errors \cite{bertinetto2020making}. Deep learning methods have enabled hierarchical information to be integrated directly into neural architectures and training objectives. Soft-label approaches smooth one-hot targets into distributions over the path from root to leaf so that intermediate representations respect the underlying taxonomy \cite{bertinetto2020making}. Mask-based designs instead impose structural constraints at the output layer: given a predicted or known parent label, binary masks zero out logits corresponding to invalid child categories, restricting predictions to taxonomically admissible descendants \cite{bouadjenek2021mask}. These strategies have shown benefits in domains such as product categorization and document tagging, where large label spaces are naturally organized into category trees, improving prediction consistency and interpretability compared to flat baselines with unconstrained outputs \cite{silla2011}.

Multi-task learning \cite{caruana1997multitask,goodfellow2016}, where models jointly optimize multiple related objectives, provides a theoretical foundation for hierarchical architectures that predict all taxonomic levels simultaneously. When multiple related tasks share representations through a common backbone, joint optimization can improve individual task performance through complementary learning signals. Recent work in biological classification has demonstrated the effectiveness of shared-backbone architectures for multi-level taxonomic prediction \cite{elhamod2022hierarchy,wittich2018}.

\subsection{Hierarchical Classification in Biology}

In biological and ecological applications, hierarchical structure is intrinsic because taxa are defined by nested morphological or genetic criteria \cite{round1990, krammer2012}. In the context of diatoms, Dimitrovski et al.\ \cite{dimitrovski2012} proposed one of the earliest hierarchical multi-label systems by combining hand-crafted shape and texture features with ensembles of predictive clustering trees to predict multiple taxonomic ranks simultaneously. Chaushevska et al.\ \cite{chaushevska2020} later incorporated CNN-based transfer-learning features into hierarchical ensembles, reporting improvements over purely hand-crafted pipelines. However, these works rely on non-end-to-end classifiers that treat hierarchy as a post-processing step rather than an architectural prior, and do not integrate hierarchical constraints into a single differentiable deep model.

Beyond diatoms, hierarchical deep models have been investigated for plant and leaf identification. Wittich \cite{wittich2018} trained CNNs to predict plant family and genus independently, demonstrating that models can generalize taxonomic knowledge from seen to unseen species. Elhamod et al.\ \cite{elhamod2022hierarchy} proposed Hierarchy-Guided Neural Networks (HGNN), which incorporate genus-level information into species-level feature learning through parallel prediction branches. Studies in this area report that enforcing taxonomic structure (either via multi-task learning across ranks or through constrained decoding) reduces implausible errors and improves the usability of automated tools for field biologists \cite{wittich2018,elhamod2022hierarchy}.

Nevertheless, to the best of our knowledge, end-to-end CNN architectures that embed taxonomic constraints \textbf{at every prediction level} have not yet been systematically evaluated for diatom classification. Existing deep-learning-based diatom systems remain predominantly flat \cite{gonzalez2025,diatomnet2024}, while existing hierarchical diatom methods rely on hand-crafted features or non-differentiable ensembles \cite{dimitrovski2012,chaushevska2020}, leaving a gap between flat CNN architectures and end-to-end hierarchical models with explicit structural constraints in this domain.



\begin{table*}[!t]
\centering
\caption{Model Architecture Summary: Prediction Targets, Head Structures, Feature Inputs, and Loss Configurations}
\label{tab:model_architecture}
\small
\begin{tabular}{llllll}
\toprule
\textbf{Model} & \textbf{Type} & \textbf{Predicts} & \textbf{Head Structure} & \textbf{Feature Input} & \textbf{Loss Weights} \\
\midrule
F-C & Flat & Class & 2-layer MLP & Backbone only & Single focal \\
F-S & Flat & Species & 2-layer MLP & Backbone only & Single focal \\
H-CO & Hierarchical & Class $\to$ Order & 2L + 3L & Backbone + parents & (1.0, 1.0) \\
H-COF & Hierarchical & Class $\to$ Family & 2L + 3L + 3L & Backbone + parents & (1.0, 1.0, 1.0) \\
H-COFG & Hierarchical & Class $\to$ Genus & 2L + 3L + 3L + 4L & Backbone + parents & (0.8, 0.9, 1.0, 1.2) \\
H-COFGS & Hierarchical & Class $\to$ Species & 2L + 3L + 3L + 4L + 4L & Backbone + parents & (0.8, 0.9, 1.0, 1.2, 1.5) \\
\bottomrule
\end{tabular}
\vspace{2mm}
\scriptsize
\begin{flushleft}
Note: Backbone is EfficientNet-B0 (1280-dim). 2L denotes [1280$\to$512$\to n$] with dropout [0.3, 0.2]. 3L denotes [input$\to$512$\to$256$\to n$] with dropout [0.3, 0.2, 0.1]. 4L denotes [input$\to$1024$\to$512$\to$256$\to n$] with dropout [0.3, 0.2, 0.2, 0.1]. All hidden layers use ReLU. Input dimension equals 1280 plus the sum of parent output dimensions. Hierarchical models use masked softmax. Focal loss uses $\alpha=0.25$, $\gamma=2.0$. Weights listed from coarse to fine.
\end{flushleft}
\end{table*}

\begin{figure*}[t]
    \centering
    \includegraphics[width=0.95\textwidth]{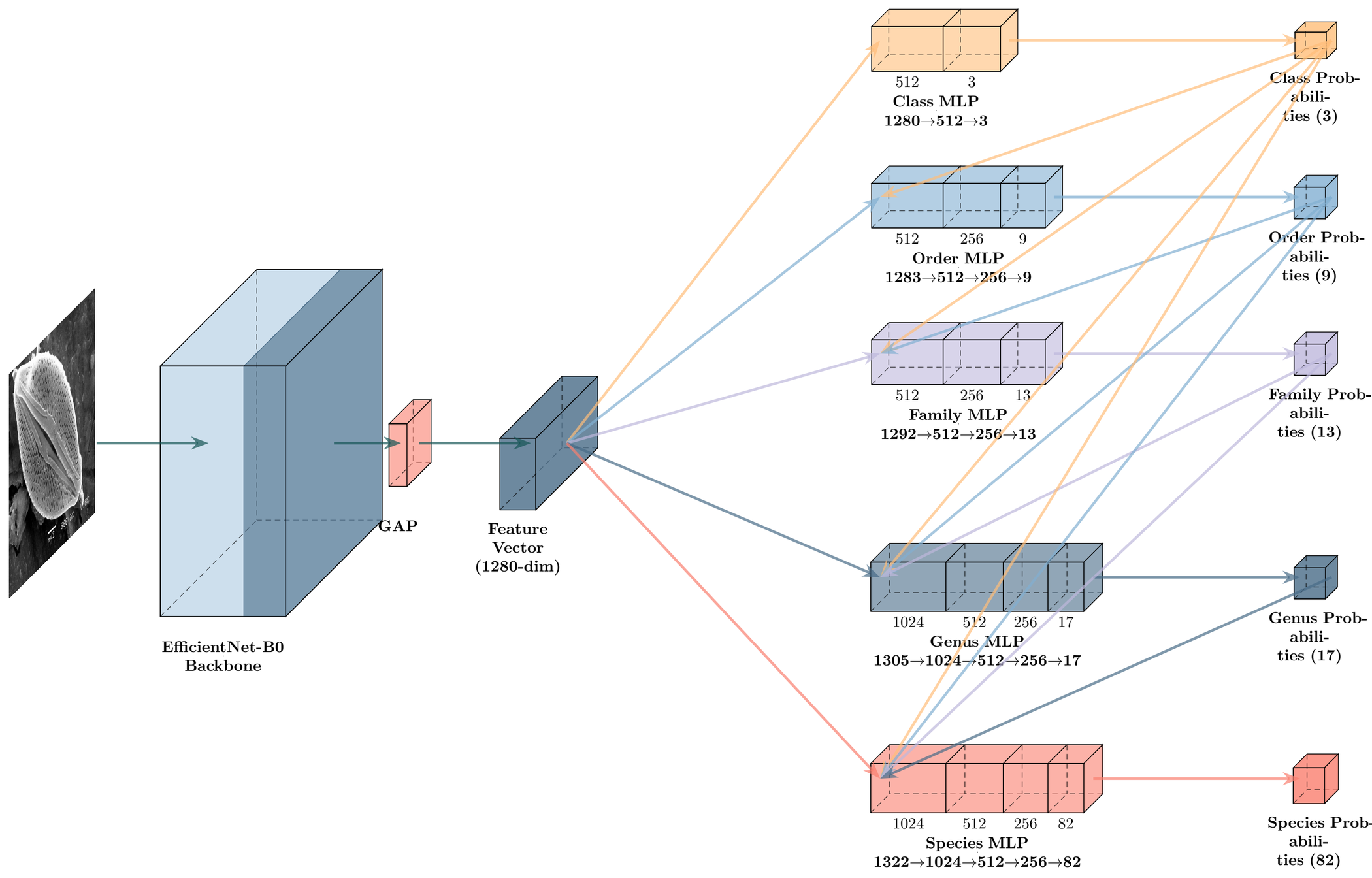}
    \caption{Hierarchical diatom classification architecture (H-COFGS). The EfficientNet-B0 backbone extracts features from input images, which are fed to five cascaded classification heads (Class, Order, Family, Genus, Species). Each head receives backbone features concatenated with probability distributions from all parent levels. Input diatom image adapted from \cite{berezovska2016diatom}, CC BY-SA 4.0.}
    \label{fig:model_architecture}
\end{figure*}

\begin{table*}[t]
\centering
\caption{Hierarchical masking example at the order level. Given predicted class Bacillariophyceae (pennate diatoms). Orders from class Bacillariophyceae receive mask value 1, while the ones from other classes are masked with $-\infty$, yielding zero probability after softmax. Softmax is computed as part of the focal loss during training.}
\label{tab:mask_example}
\begin{tabular}{llcccc}
\toprule
\textbf{Order} & \textbf{Parent Class} & \textbf{Logit} & \textbf{Mask} & \textbf{Masked Logit} & \textbf{Probability} \\
\midrule
Naviculales & Bacillariophyceae & 4.2 & 1 & 4.2 & 0.730 \\
Eunotiales & Bacillariophyceae & 2.8 & 1 & 2.8 & 0.180 \\
Cymbellales & Bacillariophyceae & 1.5 & 1 & 1.5 & 0.049 \\
Mastogloiales & Bacillariophyceae & 0.9 & 1 & 0.9 & 0.027 \\
Cocconeidales & Bacillariophyceae & 0.3 & 1 & 0.3 & 0.015 \\
\midrule
Thalassiosirales & Coscinodiscophyceae & 3.1 & 0 & $-\infty$ & 0 \\
Melosirales & Coscinodiscophyceae & 1.2 & 0 & $-\infty$ & 0 \\
Fragilariales & Fragilariophyceae & 2.5 & 0 & $-\infty$ & 0 \\
Tabellariales & Fragilariophyceae & 0.8 & 0 & $-\infty$ & 0 \\
\midrule
\multicolumn{5}{r}{\textit{Sum of valid probabilities:}} & \textbf{1.000} \\
\bottomrule
\end{tabular}
\vspace{2mm}

\raggedright
\small
\textbf{Transformation:} Logit $\xrightarrow{\text{mask}}$ Masked Logit $\xrightarrow{\text{softmax}}$ Probability. 
Invalid orders ($\text{Mask}=0$) are set to $-\infty$, so $e^{-\infty}=0$ after softmax.
Note: Thalassiosirales has a higher logit (3.1) than Cymbellales (1.5), but is masked out due to taxonomic constraints.
\end{table*}

\section{Materials and Methods}

\subsection{Dataset and Data Pipeline}

\subsubsection{Data Cleaning and Taxonomy Construction}
The initial raw dataset contains 4,881 microscope images extracted from the taxonomic reference work by Krammer and Lange-Bertalot \cite{krammer2012} for research purposes. Taxonomic labels were manually verified and annotated by research staff and students at the Yantai Institute of Coastal Zone Research, Chinese Academy of Sciences, under the supervision of Dr. Yin-chu Wang. The dataset spans 4 classes, 15 orders, 28 families, 47 genera, and 684 species, following the taxonomic system of Round et al. \cite{round1990}. 

\textit{Note: The image data is used solely for academic research and model training; no original images are published or distributed. All experimental results and visualizations in this paper use data derived from model predictions, not source images.}

We apply systematic data cleaning to correct taxonomic inconsistencies: (1) merge the rare class Mediophyceae into 
Coscinodiscophyceae \cite{round1990}, (2) remove samples with incomplete annotations at class, order, or family levels, (3) exclude species with uncertain labels containing markers such as ``sp.'', ``cf.'', or ``?'', and (4) resolve hierarchical conflicts where taxa map to multiple parents using majority voting. After cleaning, the dataset contains 3,522 samples across 3 classes, 13 orders, 23 families, 36 genera, and 608 species.

From the cleaned labels, we construct a Class $\rightarrow$ Order $\rightarrow$ Family $\rightarrow$ Genus $\rightarrow$ Species taxonomy tree, which provides parent-child mappings for hierarchical mask construction during training and inference, and enables efficient lookup for flat baselines.

\subsubsection{Progressive Minimum Sample Filtering}

To support progressive training and fair comparison, we apply a uniform threshold of 10 images per taxon at each taxonomic level, yielding four nested datasets: H-CO (3,506 samples, filters order), H-COF (3,495 samples, filters order and family), H-COFG (3,472 samples, filters order, family, and genus), and H-COFGS (1,456 samples, filters all four levels). The final filtered dataset, used by both hierarchical models H-COFGS and the flat baseline F-S, spans 3 classes, 9 orders, 13 families, 17 genera, and 82 species.

All datasets are randomly split into training (70\%), validation (15\%), and test (15\%) sets, stratified by the deepest taxonomic level of each model (e.g., class for F-C, order for H-CO) with a fixed random seed (42) to ensure reproducibility. Input images are resized to $320 \times 320$ pixels and normalized using ImageNet statistics. Data augmentation during training includes random horizontal flips, rotations within $\pm 15^\circ$, and color jittering \cite{krizhevsky2012imagenet,shorten2019survey}.

Figure~\ref{fig:taxonomy_pyramids} summarizes the taxonomy sizes and sample counts across all preprocessing stages, from the raw dataset to the progressively filtered H-COFGS subset.

\subsection{Model Architectures}

We evaluate six model variants built on EfficientNet-B0 \cite{tan2019efficientnet} pretrained on ImageNet \cite{deng2009imagenet}: two flat baselines (F-C, F-S) and four hierarchical models (H-CO, H-COF, H-COFG, H-COFGS). All models take as input preprocessed images of size $320 \times 320 \times 3$ (RGB channels), normalized using ImageNet mean and standard deviation. Table~\ref{tab:model_architecture} summarizes their architectures.

\subsubsection{Flat Models}

Flat models predict a single taxonomic level through a two-layer MLP head applied to backbone features, with labels at other levels obtained by querying the taxonomy tree.

\subsubsection{Hierarchical Models}

Hierarchical models explicitly embed taxonomic relationships by predicting labels at multiple levels in a single forward pass through cascaded classification heads (Fig.~\ref{fig:model_architecture}). Our architecture incorporates two key mechanisms: cascaded probability propagation and hierarchical mask constraints.

\textbf{Cascaded Probability Propagation.} Prior hierarchical methods either use separate CNNs at each level \cite{yan2015hdcnn} or parallel branches that predict levels independently without inter-level information flow \cite{elhamod2022hierarchy}. In contrast, our cascaded architecture explicitly propagates probability distributions from parent levels to child levels. Each head at level $\ell$ receives concatenated features from the backbone and all parent-level probability distributions (Equation~\ref{eq:feature_concat}):

\begin{equation}
\label{eq:feature_concat}
\mathbf{x}_\ell = [\mathbf{f}_{\text{backbone}}; \mathbf{p}_{\text{class}}; 
\mathbf{p}_{\text{order}}; \ldots; \mathbf{p}_{\ell-1}].
\end{equation}
This feature fusion provides a hierarchical context to each classification head, 
strengthening taxonomic consistency across levels. As illustrated in 
Fig.~\ref{fig:model_architecture}, head architectures scale with task complexity: the class head uses a two-layer MLP, intermediate levels (order, family) use three-layer MLPs, and fine-grained levels (genus, species) use four-layer architectures (detailed in Table~\ref{tab:model_architecture}).

\textbf{Hierarchical Mask Constraints.}
Another key design is the masking mechanism that enforces taxonomic consistency. 
At each level $\ell$, the model produces raw logits $\mathbf{z}_\ell$ 
transformed via masked softmax with a binary mask 
$\mathbf{m}_\ell \in \{0, 1\}^{n_\ell}$:

\begin{equation}
\label{eq:masked_softmax}
[\tilde{\mathbf{z}}_\ell]_i = \begin{cases}
[\mathbf{z}_\ell]_i & \text{if } [\mathbf{m}_\ell]_i = 1 \\
-\infty & \text{if } [\mathbf{m}_\ell]_i = 0
\end{cases}, \quad
\mathbf{p}_\ell = \text{softmax}(\tilde{\mathbf{z}}_\ell),
\end{equation}
where $\mathbf{m}\ell$ is constructed by querying the taxonomy tree based on the parent label. Table~\ref{tab:mask_example} illustrates this mechanism. Table~\ref{tab:mask_example} 
illustrates this mechanism with a concrete example from our taxonomy.

\subsection{Training Strategy}

\subsubsection{Multi-Level Loss Function}

For hierarchical models, each taxonomic level receives a separate focal 
loss \cite{lin2017focal} to address label imbalance, and the total loss 
is a weighted sum:

\begin{equation}
\label{eq:total_loss}
\mathcal{L}_{\text{total}} = \sum_{\ell \in \text{levels}} w_\ell \, \mathcal{L}_\ell^{\text{focal}},
\end{equation}
where weights increase progressively from coarse to fine levels 
(class: 0.8, order: 0.9, family: 1.0, genus: 1.2, species: 1.5) to 
emphasize harder, fine-grained predictions. Focal loss with 
$\alpha = 0.25$ and $\gamma = 2.0$ addresses label imbalance 
\cite{lin2017focal,buda2018systematic}.

\subsubsection{Training and Validation}

All models are trained using the AdamW optimizer~\cite{loshchilov2019adamw} with an initial learning rate of $5 \times 10^{-4}$, weight decay $1 \times 10^{-4}$, a ReduceLROnPlateau scheduler (decay factor 0.5, patience 5 epochs), batch size 32, a maximum of 80 epochs, and early stopping with a patience of 15 epochs based on validation performance (deepest-level weighted F1 score).

\textbf{Bidirectional Information Flow}. Standard deep learning training typically follows four steps: (1) a forward pass that maps inputs to logits, (2) loss computation based on logits and ground-truth labels, (3) a backward pass that propagates gradients through the network, and (4) parameter updates performed by an optimizer. In our hierarchical setting, the forward pass is specialized: logits are computed sequentially from coarse to fine levels, and each head receives a feature vector obtained by concatenating the shared backbone representation with all previous levels’ probability distributions (Equation~\ref{eq:feature_concat}), so that hierarchical information flows explicitly from parent to child. During loss computation, we further exploit the taxonomy by applying ground-truth-derived hierarchical masks to the logits at each level and then feeding these masked logits into a weighted focal loss (Equation~\ref{eq:total_loss}), effectively restricting every level’s candidate space to valid children of its parent and stabilizing training. The subsequent backward pass aggregates gradients from all levels and propagates them through the shared backbone, yielding bidirectional information flow: bottom-up via gradient backpropagation and top-down via the hierarchical constraints imposed during loss computation.

During validation, hierarchical models switch to greedy hierarchical prediction, which differs critically from teacher forcing: masks are constructed from predicted parent labels (from argmax at previous levels) rather than ground-truth labels. This ensures that validation performance accurately reflects inference conditions where ground-truth is unavailable, and enables proper model selection via early stopping based on realistic performance estimates.

We adopt a progressive training strategy across taxonomic depths. First, the flat class-only baseline F-C is trained from ImageNet initialization on the full cleaned dataset (3522 samples). Next, H-CO initializes its backbone from F-C and is trained on the class–order dataset H-CO (3506 samples), which is filtered to satisfy the minimum-sample threshold at the order level. Similarly, H-COF initializes its backbone from H-CO and trains on the H-COF dataset (3495 samples), H-COFG initializes from H-COF and trains on H-COFG (3472 samples), and H-COFGS initializes from H-COFG and trains on H-COFGS (1456 samples), each using its own filtered dataset that enforces the minimum-sample threshold at its deepest taxonomic level.

\begin{figure}[!t]
\centering
\includegraphics[width=\columnwidth]{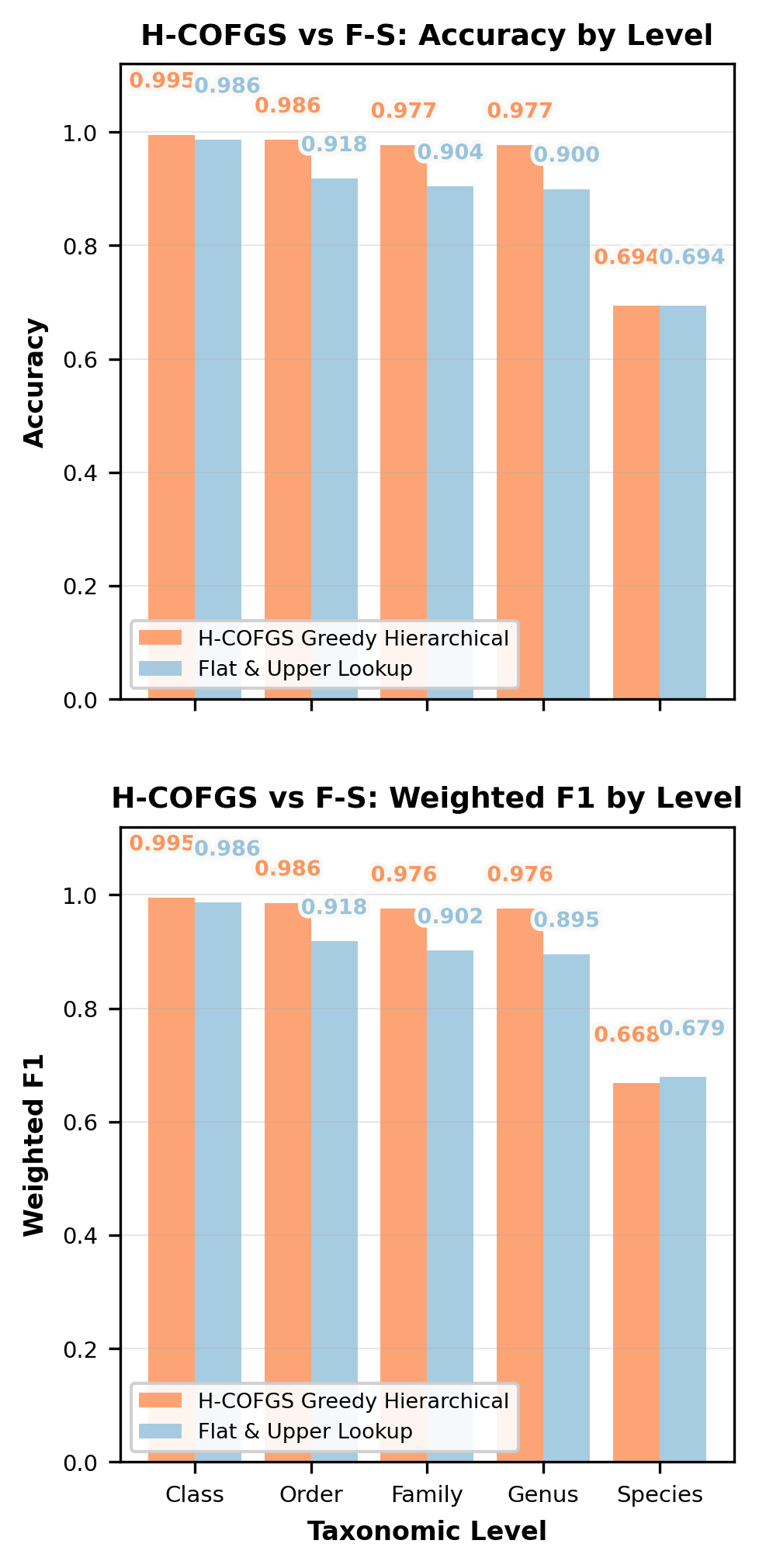}
\caption{Performance comparison between H-COFGS (hierarchical) and F-S (flat with taxonomy tree lookup) across all taxonomic levels on the final filtered dataset of 1,456 images. Both models achieve comparable species-level performance, but H-COFGS substantially outperforms F-S at all upper levels.}
\label{fig:hcofgs_vs_fs}
\end{figure}

\subsubsection{Inference}

During test inference, flat models predict a single taxonomic level directly, with labels at other levels obtained by querying the taxonomy tree. Hierarchical models use the same greedy hierarchical prediction as validation: logits are computed sequentially (Equation~\ref{eq:feature_concat}), probability distributions are computed via softmax, and we greedily select the category with the highest probability (argmax) at each level, using the predicted index from the previous level to mask the current level's probability distribution via Equation~\ref{eq:masked_softmax}, ensuring biological consistency:

\begin{align}
\hat{y}_{\text{class}} &= \arg\max(\text{softmax}(\mathbf{z}_{\text{class}})), \\
\mathbf{m}_{\text{order}}^{\text{infer}} &= \text{get\_mask}(\hat{y}_{\text{class}}), \\
\hat{y}_{\text{order}} &= \arg\max(\text{softmax}(\mathbf{z}_{\text{order}} \odot \mathbf{m}_{\text{order}}^{\text{infer}})).
\end{align}

This procedure guarantees taxonomic consistency by construction. Alternative inference strategies, including beam search and level-wise argmax (without hierarchical constraints), are evaluated in the appendix.


\section{Experimental Results}

\subsection{Hierarchical vs Flat Performance Comparison}

Figure~\ref{fig:hcofgs_vs_fs} presents a direct comparison between the hierarchical model H-COFGS and the flat baseline F-S on the same test set (219 samples, 82 species). Both models achieve comparable species-level performance: identical accuracy (69.4\%) with F-S slightly outperforming H-COFGS in weighted F1 (0.679 vs 0.668). At upper taxonomic levels, H-COFGS substantially outperforms F-S: class accuracy is 99.5\% versus 98.6\%, while order, family, and genus accuracies all exceed 97\% compared to 90--92\% for the flat baseline. Weighted F1-scores at upper levels exhibit similar patterns.

\subsection{Error Robustness and Taxonomic Distance Analysis}


\begin{figure}[!t]
\centering
\includegraphics[width=\columnwidth]{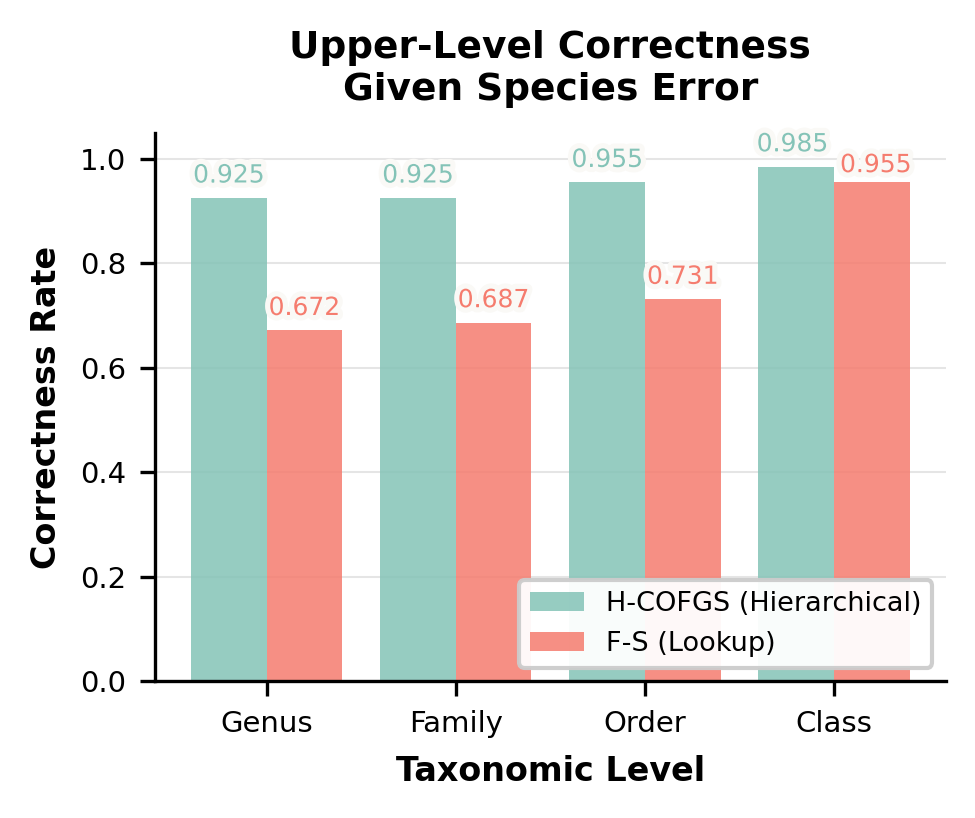}
\caption{Error propagation comparison between hierarchical (H-COFGS) and flat (F-S) models. When both models misclassify species, H-COFGS maintains substantially higher accuracy at all ancestor levels.}
\label{fig:error_propagation}
\end{figure}

We analyze error robustness by examining how errors propagate across taxonomic levels. This analysis is relevant for biological applications where taxonomically local errors (e.g., confusing species within the same genus) are far less consequential than errors that cross family or class boundaries.

Figure~\ref{fig:error_propagation} presents upper-level correctness conditioned on species-level errors. When species are misclassified, H-COFGS maintains substantially higher correctness at all ancestor levels.


\begin{table}[!t]
\centering
\caption{Taxonomic Distance Analysis on Species-Level Errors}
\label{tab:error_distance}
\small
\begin{tabular}{lcc}
\toprule
\textbf{Taxonomic Distance} & \textbf{H-COFGS} & \textbf{F-S} \\
\midrule
Distance = 0 (correct species) & 152 & 152 \\
Distance = 1 (same genus)      & 62  & 45  \\
Distance = 2 (same family)     & 0   & 1   \\
Distance = 3 (same order)      & 2   & 3   \\
Distance = 4 (same class)      & 2   & 15  \\
Distance $\geq$ 5 (cross class) & 1  & 3   \\
\midrule
Total test samples             & 219 & 219 \\
Species errors                 & 67  & 67  \\
\midrule
\multicolumn{3}{l}{\textit{Share of 67 species errors at each distance}} \\
\midrule
Distance = 1 (same genus)      & 92.5\% & 67.2\% \\
Distance = 2 (same family)     & 0.0\%  & 1.5\%  \\
Distance = 3 (same order)      & 3.0\%  & 4.5\%  \\
Distance = 4 (same class)      & 3.0\%  & 22.4\% \\
Distance $\geq$ 5 (cross class) & 1.5\% & 4.5\%  \\
\midrule
Mean distance (all samples)    & 0.370  & 0.598  \\
Mean distance (errors only)    & 1.209  & 1.955  \\
Std dev (errors only)          & 0.687  & 1.423  \\
\midrule
Error severity reduction       & \multicolumn{2}{c}{38.2\%} \\
\bottomrule
\end{tabular}
\vspace{2mm}
\scriptsize
\begin{flushleft}
Note: Taxonomic distance is the number of edges in the taxonomy tree between predicted and true species. Distances $\geq 1$ are normalized by the 67 species-level errors for each model.
\end{flushleft}
\end{table}

Table~\ref{tab:error_distance} quantifies error locality through taxonomic distance analysis, defined as the number of edges in the taxonomy tree between the predicted and true species. H-COFGS produces 92.5\% of errors at distance 1 (same genus) compared to 67.2\% for F-S. The mean taxonomic distance for errors is 1.209 for H-COFGS versus 1.955 for F-S, representing a 38.2\% reduction. 

\subsection{Progressive Training Analysis}

\begin{figure*}[!t]
\centering
\includegraphics[width=\textwidth]{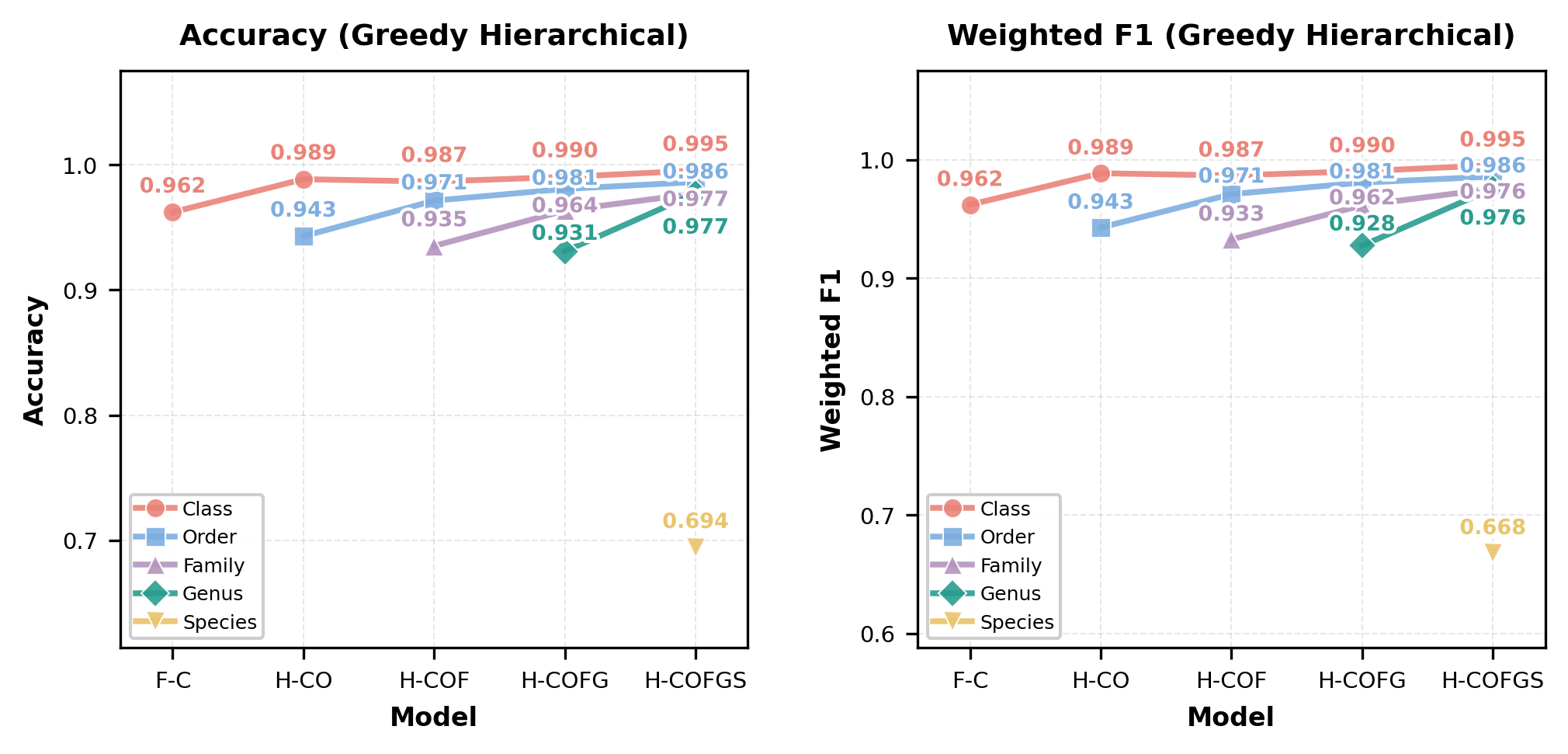}
\caption{Performance evolution across progressive hierarchical models. As deeper taxonomic levels are added (H-CO $\to$ H-COF $\to$ H-COFG $\to$ H-COFGS), upper-level performance shows a general upward trend.}
\label{fig:progressive_training}
\end{figure*}

Figure~\ref{fig:progressive_training} presents results from progressive training experiments, where we sequentially add taxonomic levels (H-CO $\to$ H-COF $\to$ H-COFG $\to$ H-COFGS). Class accuracy provides a representative example: it improves from 96.2\% (F-C) to 99.5\% (H-COFGS), despite a slight drop at H-COF. Order, family, and genus accuracies exhibit similar upward trends.

\section{Discussion}

\subsection{Why Does Hierarchical Architecture Improve Upper-Level Performance?}

Figure~\ref{fig:hcofgs_vs_fs} shows that H-COFGS matches F-S at the species level but substantially outperforms it at upper levels. Why does the hierarchical model achieve better upper-level performance without sacrificing species-level accuracy?

We attribute this to direct supervision at each taxonomic level during training. Flat models receive supervision only at the species level; upper-level labels are derived deterministically through tree lookup and receive no gradient signal during training. The model learns features optimized solely for 82-way species discrimination. In contrast, hierarchical models receive explicit loss signals at class, order, family, genus, and species simultaneously. This creates a multi-objective optimization problem where the shared backbone must learn representations that balance discriminability across multiple taxonomic scales, which is from macro-scale traits distinguishing classes to micro-scale details distinguishing species.

This multi-level supervision defines a hierarchical multi-task learning objective that acts as an implicit regularizer, encouraging the backbone to learn more general and robust representations~\cite{caruana1997multitask}. In practice, this allows the model to maintain strong species-level performance while simultaneously improving discriminability at coarser taxonomic levels.

\subsection{The Role of Bidirectional Information Flow}

The hierarchical model implements bidirectional information flow through two mechanisms during training:

\begin{enumerate}
    \item \textbf{Top-down hierarchical constraints}: Masked softmax (Equation~\ref{eq:masked_softmax}) constructed from ground-truth parent labels constrains the prediction space at each level during training. This ensures that the model learns to respect taxonomic structure and reduces the search space for fine-grained predictions.
    
    \item \textbf{Bottom-up gradient backpropagation}: Losses from all five taxonomic levels (Equation~\ref{eq:total_loss}) propagate gradients backward through the shared backbone (Figure~\ref{fig:model_architecture}). Fine-grained supervision signals from species and genus levels flow back to update early convolutional layers, refining low- and mid-level feature representations that also serve coarser predictions.
\end{enumerate}

Progressive training experiments (Figure~\ref{fig:progressive_training}) provide evidence for this mechanism: class accuracy improves from 96.2\% (F-C) to 99.5\% (H-COFGS) as deeper taxonomic levels are added. This upward trend suggests that fine-grained supervision refines the backbone's feature representations, benefiting all levels, including the coarsest.

\subsection{Error Locality and Biological Interpretability}

From a practical standpoint, error locality is crucial for biological applications. Confusing two species within the same genus (e.g., \textit{Navicula radiosa} vs.\ \textit{Navicula cryptotenella}) is far less consequential than confusing species across different classes (e.g., a centric diatom misclassified as a pennate diatom). The former may reflect genuine morphological overlap or ambiguous specimens, whereas the latter indicates fundamental misunderstanding of diatom morphology.

Table~\ref{tab:error_distance} shows that H-COFGS concentrates 92.5\% of its species-level errors within the same genus (distance = 1), compared to only 67.2\% for F-S. The flat baseline produces significantly more errors at distance 4 (same class but different order or family): 22.4\% versus 3.0\% for H-COFGS. This difference has meaningful implications: hierarchical predictions are more aligned with biological structure, producing mistakes that remain taxonomically local and thus easier for domain experts to interpret and correct.

The mean taxonomic distance (1.209 vs. 1.955) serves as a useful metric for model quality beyond top-1 accuracy. Two models with identical species-level accuracy may differ substantially in the locality of their errors, and the hierarchical architecture provides a mechanism that favors biologically plausible mistakes.

\subsection{Limitations}

Our study has several limitations. First, the final filtered dataset H-COFGS contains only 1,456 images across 82 species. While this ensures sufficient samples per taxon (minimum 10), it is small by modern deep learning standards. Larger datasets covering more genera and species would provide a stronger test of the hierarchical architecture's generalization capability.

Second, we evaluate only one backbone architecture (EfficientNet-B0). Other architectures, such as ResNet, Vision Transformers, or ConvNeXt, may exhibit different behavior, and the relative benefits of hierarchical prediction may vary with backbone capacity and inductive biases.

Third, the dataset originates from a specific imaging protocol and geographic region. Domain shift—differences in microscope settings, staining procedures, or regional species distributions—remains a challenge for real-world deployment and requires evaluation on external datasets.

\subsection{Future Directions}

Several directions could extend this work. Larger datasets covering more taxonomic groups would enable a more robust evaluation of hierarchical architectures. Incorporating semi-supervised or self-supervised learning could leverage abundant unlabeled diatom imagery to improve feature representations.

Integrating uncertainty estimation into hierarchical predictions would enhance trustworthiness: the model could abstain from fine-grained predictions when uncertain and fall back to genus- or family-level labels. Bayesian approaches or ensemble methods could provide calibrated confidence scores at each taxonomic level.

Extending the architecture to multi-modal inputs (e.g., combining bright-field and phase-contrast images, or integrating environmental metadata such as pH and conductivity) could improve accuracy and robustness. Hierarchical structures are naturally suited to multi-modal fusion, as different modalities may provide stronger signals at different taxonomic levels.

Finally, while this work focuses on diatoms, the hierarchical framework is broadly applicable to other biological classification tasks with nested taxonomic structure, such as plant identification, insect classification, or microbial ecology. Evaluating the approach across diverse domains would clarify the generality of the observed benefits.

\section{Conclusion}

This work examined whether explicitly encoding taxonomic hierarchy in deep learning architectures improves diatom classification compared with flat single-level prediction. We introduced a hierarchical convolutional network with five cascaded heads that jointly predict class, order, family, genus, and species via cascaded probability propagation and taxonomy-derived masks that restrict each level to valid descendants during both training and greedy hierarchical inference. On a filtered dataset of 1,456 images spanning 82 species, the hierarchical model matches flat baselines at the species level while substantially outperforming them at all upper taxonomic levels and producing more taxonomically local errors when fine-grained predictions fail.

Progressive training experiments suggest that hierarchical masks provide top-down constraints on the prediction space, while losses at all taxonomic levels drive bottom-up gradients through the shared backbone, yielding a hierarchical multi-task objective that improves both accuracy and biological interpretability. These results indicate that hierarchical architectures offer a principled way to embed domain structure into deep learning models for diatom classification, and motivate future work on larger datasets, richer inference and uncertainty estimation, and applications to other taxa with nested taxonomic structure.  

\section*{Acknowledgment}
I would like to thank Dr. Yin-chu Wang for providing the diatom dataset and for insightful guidance on biological taxonomy throughout this project. In line with course policy, AI-assisted tools were used for brainstorming, grammar and IEEE style checking, and debugging code snippets. All experimental design, model implementation, written content, data analysis, and conclusions were performed and verified by me.

\appendices

\section{Alternative Inference Strategies}

The main text uses greedy hierarchical prediction during inference: at each taxonomic level, we select the category with the highest probability (argmax) and use it to construct the mask for the next level. This procedure guarantees taxonomic consistency but may propagate errors from upper to lower levels. Here we evaluate two alternative strategies:

\begin{enumerate}
    \item \textbf{Level-wise argmax without hierarchical constraints}: At each level, select the category with the highest probability independently, ignoring taxonomic consistency. This may produce biologically implausible predictions (e.g., a genus not belonging to the predicted family).
    
    \item \textbf{Beam search}: Maintain top-$k$ candidates at each level and propagate them to the next level, selecting the final prediction by maximizing the product of probabilities across all levels. This explores a larger hypothesis space than greedy decoding.
\end{enumerate}

\begin{figure}[!t]
\centering
\includegraphics[width=\columnwidth]{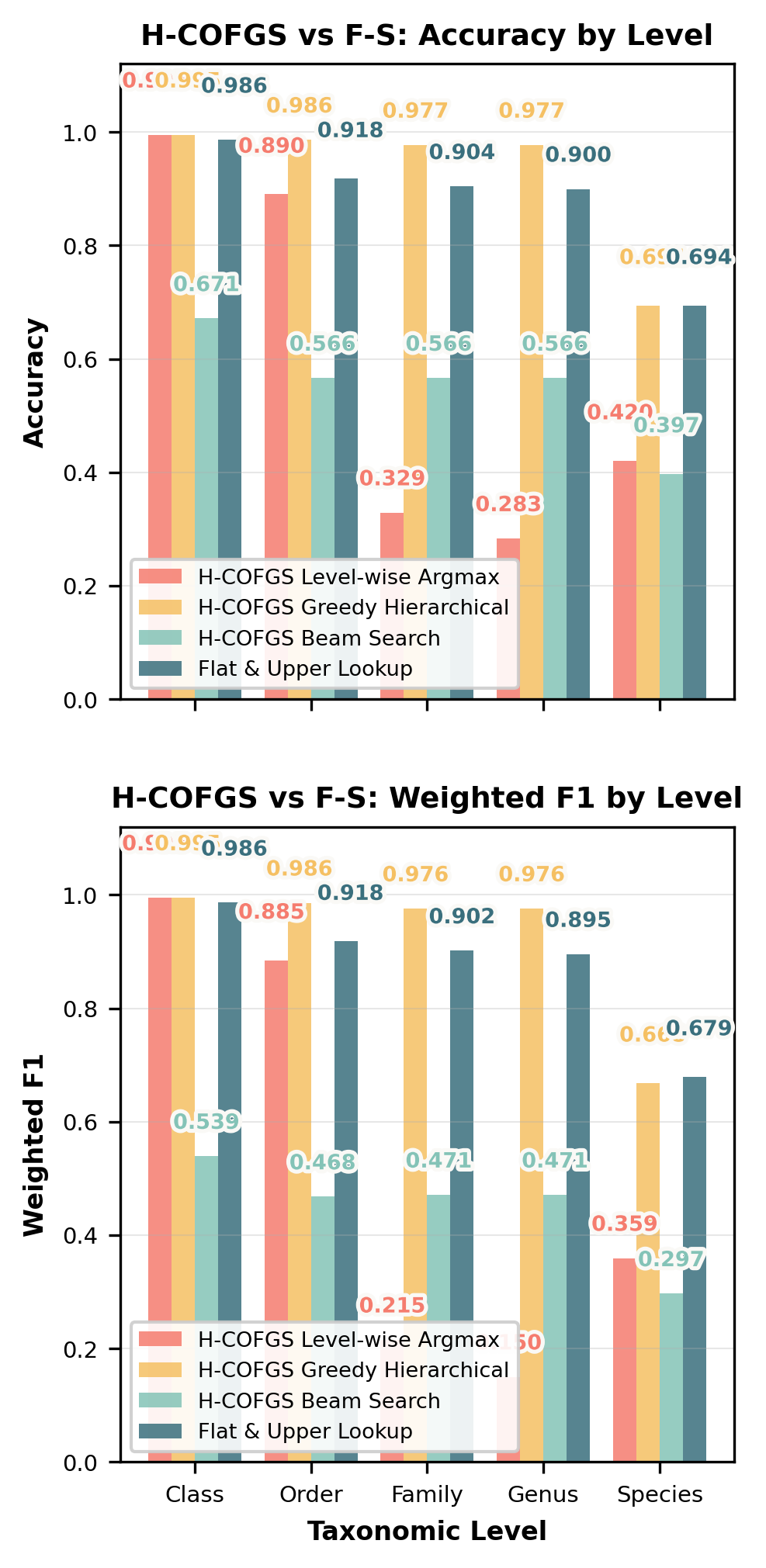}
\caption{Comparison of inference strategies across taxonomic levels. Greedy hierarchical prediction (with masked constraints) achieves the best balance between accuracy and taxonomic consistency.}
\label{fig:appendix_inference}
\end{figure}

Figure~6 compares four prediction strategies on the H-COFGS test set: H-COFGS level-wise argmax without constraints, H-COFGS greedy hierarchical decoding, H-COFGS beam search with width $k=3$, and the flat F-S baseline with bottom-up lookup for upper levels. At the species level, greedy hierarchical decoding achieves 69.4\% accuracy, compared to 67.9\% for F-S, 42.0\% for level-wise argmax, and 39.7\% for beam search.

\bibliographystyle{IEEEtran}
\bibliography{references}

\end{document}